\def\year{2021}\relax
\documentclass[letterpaper]{article} 
\usepackage{aaai21}  
\usepackage{times}  
\usepackage{helvet} 
\usepackage{courier}  
\usepackage[hyphens]{url}  
\usepackage{graphicx} 
\usepackage{natbib}  
\usepackage{caption} 
\usepackage{amsmath, amsfonts}
\usepackage{booktabs}
\usepackage[table,xcdraw]{xcolor}

\title{A Neuro-Symbolic Method for Solving Differential and Functional Equations}
\author {
    Maysum Panju,\textsuperscript{\rm 1}
        Ali Ghodsi \textsuperscript{\rm 2}
          \\
}
\affiliations {
    Department of Statistics and Actuarial Science \\
    University of Waterloo
}

\begin{document}

\maketitle

\begin{abstract}
When neural networks are used to solve differential equations, they usually produce solutions in the form of black-box functions that are not directly mathematically interpretable. We introduce a method for generating symbolic expressions to solve differential equations while leveraging deep learning training methods. Unlike existing methods, our system does not require learning a language model over symbolic mathematics, making it scalable, compact, and easily adaptable for a variety of tasks and configurations. As part of the method, we propose a novel neural architecture for learning mathematical expressions to optimize a customizable objective. The system is designed to always return a valid symbolic formula, generating a useful approximation when an exact analytic solution to a differential equation is not or cannot be found. We demonstrate through examples how our method can be applied on a number of differential equations, often obtaining symbolic approximations that are useful or insightful. Furthermore, we show how the system can be effortlessly generalized to find symbolic solutions to other mathematical tasks, including integration and functional equations.
\end{abstract}

\section{Introduction}

Differential equations play a fundamental role in describing the physical laws that govern the natural world, and have been the subject of much study for centuries. 
More recently, artificial neural networks have gained popularity as they have been shown to demonstrate superior performance in a wide variety of tasks. Supported by the back-propagation techniques that make deep learning possible, there are few fields remaining in which neural networks have not made a contribution.
Naturally, neural networks have been applied to solving differential equations, with promising results.

Many papers have shown the ease with which neural networks can model solutions to previously unmanageable differential equations. The function satisfying the differential equation, then, is not given in the form of a closed-form mathematical expression, but a trained neural network model that makes up for its lack of interpretability with its highly accurate fit.

Perhaps it is the success that neural networks have had with this approach that has prevented progress in using deep learning to solve differential equations in a different way: obtaining symbolic solutions in the form of readable, mathematical expressions.

 The recent paper on Deep Learning for Symbolic Mathematics \cite{lample2019deep} is one of the rare contributions that tries to take on this task. In that work, the authors demonstrate how using methods in deep learning can achieve state-of-the-art performance in solving differential equations and integration problems symbolically. Their method employs deep learning in an indirect way: rather than using deep learning to discover the solutions to differential equations themselves, their method involves constructing and making use of a language model that has been trained to ``understand'' the language of symbolic mathematics. Thus, the differential equations are not solved by using deep learning itself, but by means of a pre-trained  model that had made use of deep learning as it was set up.

In this work, we introduce a method that aims to use deep learning to solve problems in symbolic mathematics directly. We present a framework for solving differential equations that explores the space of candidate mathematical functions and uses deep learning techniques to identify the optimal solution for the problem. 

As part of this system, we include the design for a neural-based symbolic function learner (SFL) that is able to produce symbolic mathematical expressions that minimize a specified cost function. The key requirement of an SFL is that, given a class of symbolic functions $F$ and a measure of cost on functions $L:F\rightarrow \mathbb{R}$, the SFL is able to discover
\begin{equation}
    \arg\min_{f \in F} L(f).
    \label{SFL_min}
\end{equation} 

The SFL we present in this work seeks the minimizer of Equation~\ref{SFL_min} using a purely gradient-based search, allowing it to take advantage of all of the training techniques that neural networks have to offer.

As Lample and Charton admit in their paper, neural networks are rarely seen to solve problems in symbolic mathematics because neural networks are, generally, poor at dealing with symbolic mathematics. We demonstrate that this weakness is not insurmountable. While our method may not compete with state-of-the-art algorithms for automated solving of highly complex differential equations, our proposed system has the advantage of being able to produce symbolic approximate solutions for every problem that it does not or cannot solve. These are symbolic expressions that may not necessarily solve the differential equations exactly, but whose values resemble those of the true solutions over a given interval.

Symbolic approximation functions are useful because they allow insight to the shape of a solution even when the solution itself cannot be found. Furthermore, they allow us to see which functions in the function space $F$ are closest to the true solution when the true solution itself does not lie in $F$. This is particularly useful in cases where the solution to a differential equation cannot be represented in terms of elementary functions.

In the sections that follow, we outline the core of our framework and the architecture of our SFL model, and enumerate some of the unique advantages of our system. We then show several experiments that demonstrate the power of our method and the utility of symbolic approximation functions. We are able to use our method to determine symbolic functions that approximately solve a number of members in the family of Lane-Emden equations, several of which have eluded closed-form solutions. We also show how our framework can be used to solve integration problems, and generate symbolic approximate integrals to a number of functions with non-analytic antiderivatives.


\section{Related Work}

There are countless papers describing methods and applications of using neural networks for solving differential equations. Some of the interesting contributions made using this approach include \cite{meade1994solution, lagaris1998artificial, parisi2003solving, malik2020learning}. A genetic-based approach that makes use of neural networks is \cite{tsoulos2009solving}.

The main work that uses deep learning for symbolic mathematics is given in  \cite{lample2019deep}. The authors use a transformer network and an intelligently constructed training dataset to learn a language model over the language of symbolic mathematics. They treat differential equations and integration problems as input to a sequence-to-sequence network, trusting the language model to have learned the rules of mathematics after observing tens of millions of valid examples.  This approach has worked so well that they are able to integrate expressions that even Mathematica cannot. 

As pointed out in \cite{davis2019use}, this model is ``like the worst possible student in a calculus class'': it is able to work with mathematical symbols only because of an abundance of repeated exposure, and not because of any inherent understanding or appreciation for any of the concepts actually represented.

In contrast, the model we will now propose is like a student that has not studied the techniques used in the course, but tackles each calculus problem as a brand-new challenge, discovering and improving solutions in a trial-and-error based method. It may not be the best student, but it is certainly more dedicated and  adaptable than the worst.

\section{Method}

The proposed system contains two components: (1) an equation-solving wrapper and (2) an algorithm to generate symbolic functions that fits in the system as a submodule. We explain both segments in the sections that follow.

\subsection{Equation Solver System}

A general form for representing nonlinear, ordinary differential equations is 
\begin{equation}
    g(x, y, y', y'', \ldots) = 0.
    \label{g_defn}
\end{equation}

The solution to this system is the mathematical function $f(x)$ that minimizes the expected error
\begin{equation}
    L_1(f) = \mathbb{E}\left[\| g(x, f(x), f'(x), f''(x), \ldots) \|^2\right]
\end{equation}
where $x$ is distributed over the desired domain of $f$.

In some cases, the differential equation may be constrained by initial values for the function or any of its derivatives, $\{(x_i, y_i^{(n_i)})\}$. Frequently, these initial values represent boundary conditions that restrict the choices for the solution to the differential equation. In the present framework, we do not impose any requirement on whether these constraining points lie on the boundary of the domain or otherwise. 

These constraints may be attained by a function $f(x)$ that minimizes the error
\begin{equation}
    L_2(f) = \sum_i \left\| f^{(n_i)}(x_i) - y_i^{(n_i)} \right\|^2.
\end{equation}

The solution to the differential equation can be therefore found by finding the function $f$ that minimizes the error 
\begin{equation}
\label{loss_eqn}
Err = L_1(f) + \lambda L_2(f). 
\end{equation}

If $f$ belongs to a class of functions that can be learned using methods based on gradient descent, then the differential equation can be solved by standard deep learning techniques, such as back-propagation. This is how neural networks are typically used for solving differential equations.

However, if $f$ is assumed to be in the form of a symbolic mathematical expression, then in order to use Equation~\ref{loss_eqn} to learn the function via deep learning techniques, there must be a method for generating symbolic expressions that minimize a given cost function. This is the task that we aim to address using our symbolic function learner, described next.






\subsection{Symbolic Function Learner (SFL)}

\begin{figure*}[t]
\centering
\includegraphics[width=0.95\textwidth]{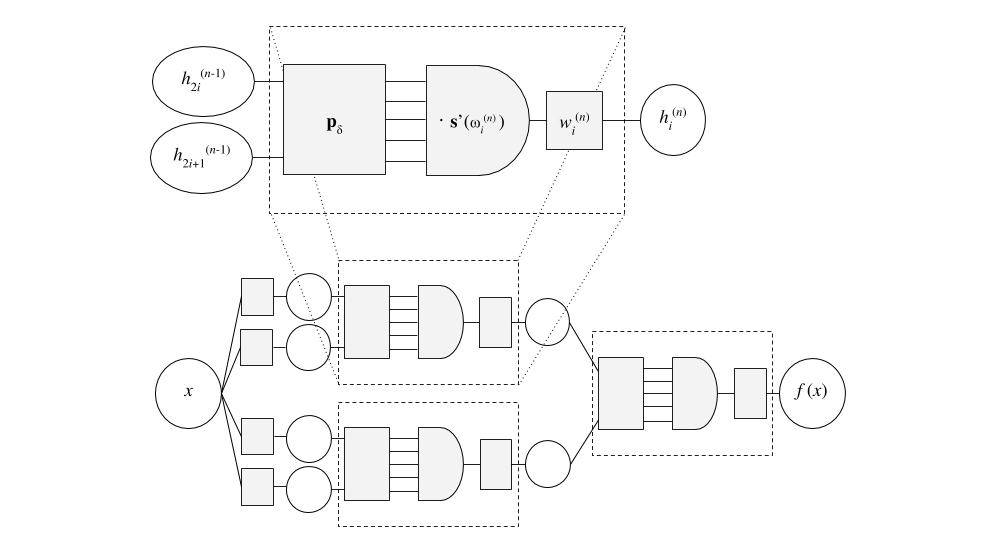} 
\caption{A schematic diagram depicting the structure of the proposed symbolic function learner (SFL). \textbf{Top:} The neural representation of a single operator node in the symbolic parse tree. The operator function $\textbf{p}_\delta$ applies all available operators on the two child node values it takes as input. The discretized softmax function $\textbf{s}'$ acts as a gate, filtering out all but one of these operators, determined by learnable weight $\omega$. This output is then scaled by learnable weight $w$. (Note: bias scalars $b$ are omitted from the diagram to save space.) \textbf{Bottom:} A series of operator nodes are assembled in a balanced binary tree, which altogether represents a single, well-formed symbolic mathematical exprsession $f$.  }
\label{architecture}
\end{figure*}

We present a neural-based model for learning symbolic mathematical expressions that can be trained to optimize a given cost function. 

Our model is based on the intuition that mathematical expressions can be expressed as syntactic parse trees, where interior nodes represent unary or binary operators that take their child nodes as input, and leaf nodes represent numerical quantities in the form of variables, constants, or a mixture of both. Every parse tree represents a single mathematical expression, although the converse does not hold.

The structure of these trees can be partially standardised by ensuring that the tree is perfectly binary (i.e. every interior node has exactly two children) by defining a way for unary operators to combine two values into a single input argument. Furthermore, we can ensure the tree is balanced (i.e. subtrees rooted at every node on any given level of the tree all have the same depth) by allowing an identity operator that can be used to add levels within the tree without changing the value of the represented expression. 

Our model can be summarized as follows: given a set of allowable operators and a maximum tree depth, we use learnable weights to determine what choice of operator will be placed in each of the interior nodes of the tree. Thus, given any present values of weights, a tree configuration can be interpreted as a mathematical expression that can be examined and evaluated. The weights are updated using back-propagation in order to optimize how well the resulting mathematical expression fits the given dataset. A more precise description of the model follows below.


Let $U$ be a list of allowable unary operators $[u_1, \ldots, u_r]$ that map $\mathbb{R}$ to $\mathbb{R}$, and let $V$ be a list of binary operators $[v_{r+1}, \ldots, v_k]$ that map $\mathbb{R}^2$ to $\mathbb{R}$, for a total of $k$ allowable operators. We define the ``operate'' function $\textbf{p}_\delta: \mathbb{R}^2 \rightarrow \mathbb{R}^k$ by 
\begin{eqnarray*}
\textbf{p}(x_1, x_2) &=& [ u_1(x_1 + \delta x_2), \ldots, u_r(x_1 + \delta x_2), \\
&& v_{r+1}(x_1, x_2), \ldots, v_k(x_1, x_2) ].
\end{eqnarray*}

Here, $\delta \in \{0, 1\}$ is a binary hyperparameter that determines whether or not to include both inputs to \textbf{p} when computing unary operators. By setting  $\delta = 1$, the space of learnable functions for a given tree depth becomes richer, and the resulting mathematical expression may be more complex. When $\delta = 0$, both the space of candidate functions and the typical length of expressions are reduced. We have observed that it is useful to let $\delta=1$ when the number of layers in the tree is small, and to let $\delta=0$ when the number of layers is large.

As an example of the operate function, if $U = [id, \sin, \sqrt{}]$, $V =  [+, \times]$, and $\delta = 0$, then $\textbf{p}(x_1, x_2) = [x_1, \sin(x_1), \sqrt{x_1}, x_1 + x_2, x_1x_2 ]. $ Note that we use $id$ to refer to the identity function that returns its input value unchanged.

In each interior node of the parse tree, the $\textbf{p}_\delta$ function is computed, obtaining the values of all possible operators taking in the values of the node's two children as input. In order to interpret the parse tree as a mathematical expression, the node should only represent one of these operators. We use a gate variable to determine which operator's output will be allowed to pass on through the node.

More precisely, let $\omega$ be a learnable weight vector in $\mathbb{R}^k$. Denote by \textbf{s} the softmax function, so that
$$ \textbf{s}(\omega) = \left[\frac{e^{\omega_i}} {\sum_{j=1}^k e^{\omega_j}} \right]_{i=1}^k.$$

Then $ \textbf{s}(\omega)$ is a vector in $\mathbb{R}^k$ with nonnegative entries summing to 1, and taking the dot product of this vector with the output of $\textbf{p}_\delta$ gives a convex linear combination of the outputs over all operators allowed by $\textbf{p}_\delta$, giving most weight to the operator at the index corresponding to the largest entry of $\omega$. In this way, the choice of operator to be active at a given node can be learned by updating the learnable weight vector $\omega$. 

The softmax gate does not entirely filter out the contributions of all but one operator in the output of $\textbf{p}$. To correct this, we adjust the output of $\textbf{s}(\omega)$ to become $\textbf{s}'(\omega)$, where the entry corresponding to the largest value of $\omega$ is 1 and all other entries are 0. This operation can be performed using functions that are differentiable almost everywhere (such as by dividing $\textbf{s}(\omega)$ by its maximum entry and passing the result through by a narrow hump function centred at 1), preserving the differentiability of our model that allows gradient-based training methods. 





Now that we have described the framework of a single node in the parse tree, we can take a look at the structure of the tree as a whole.

Let $m$ represent the number of layers in the parse tree. Note that the number of layers determines how complex the resulting mathematical expression is able to become.

For $i = 0, \ldots, 2^{m}-1$, define 
$$h_i^{(0)} = w_i^{(0)} x + b_i^{(0)}$$
where $x$ is the name of the variable in the mathematical expression being constructed, and $w_i^{(0)}$ and $b_i^{(0)}$ are learnable scalar parameters.  This represents the lowest layer of the tree, consisting of leaf nodes that denote numerical quantities.

For each subsequent layer of the tree $n = 1, \ldots, m$, recursively define the value of each node in the tree as
$$h_i^{(n)} = w_i^{(n)} \left(\textbf{s}'\left(\omega_i^{(n)}\right) \cdot \textbf{p}_\delta\left(h_{2i}^{(n-1)}, h_{2i+1}^{(n-1)})\right)  \right) + b_i^{(n)}$$
for each $i = 0, \ldots, 2^{m-n-1}$. Here, each $\omega_i^{(n)}$ is a learnable weight vector in $\mathbb{R}^k$, and $w_i^{(n)}$ and $b_i^{(n)}$ are learnable scalar weights as before.
 
When $n = m$, we get $h_0^{(m)}$ representing the value at the root node of the tree, which is the value of the resulting mathematical expression when given the input $x$. It is by using this value (obtained as a differentiable function of $x$ and all learnable weights in the SFL) that we are able to train SFL using the cost function in Equation~\ref{loss_eqn}. 

The SFL returns the symbolic expression obtained by interpreting the interior tree nodes as operators determined by the weights in each $\omega$, subjecting each node to the affine transformations defined by its corresponding $w$ and $b$.

\section{Strengths and Advantages}
 
Our proposed equation solving framework possesses the following strengths:

\subsubsection{Ability to Approximate.} Most automated systems for solving differential equations or performing integration are unable to produce a symbolic expression when given a problem that is outside of their capacity to solve \cite{moses1971symbolic}. In particular, there are many differential equations and integration problems that are known to have no closed-form analytic solution, which would be guaranteed to fail in these systems \cite{kragler2009mathematica}. 

Our framework is able to provide a symbolic expression as output in every single instance. Since the SFL is performing an optimization search based on reducing an objective value, it does not distinguish between problems that are solvable or not, and will always return the expression found during the search process that attains the best objective score. 

A powerful advantage of this fact is that our system can be used to generate optimal approximations for functions under user-imposed restrictions on expression complexity. Furthermore, our system excels at obtaining symbolic approximations for functions that cannot be expressed in terms of elementary functions at all, as will be shown in the Experiments section.

\subsubsection{Modularity.} The framework is modular and can be easily adapted as models for generating symbolic functions improve. Although we present one model for an SFL in this work, the framework can easily be applied using any other method for obtaining symbolic mathematical expressions that optimize a given cost function. 

In particular, many symbolic regression algorithms are designed to find the symbolic expression that minimizes the mean square error (or some other fitness metric) over a given dataset. In some of these approaches, the regressor is performing an optimization problem over the search space of symbolic functions. This is the case for some genetic evolution based strategies, including Eureqa \cite{schmidt2009distilling}, as well as methods that involve neural networks, including \cite{sahoo2018learning} and \cite{udrescu2020ai}. Our framework provides an excellent alternative application where these same models can be used, with little adjustment, for a different task.

\subsubsection{No Dependency on Language Model.}
The state-of-the-art method for using deep learning to solve problems in symbolic mathematics is given by Lample and Charton \cite{lample2019deep}. Their technique (which we will call ``LC''), while capable of producing outstanding results, necessarily requires learning a language model for symbolic expressions before it can be used. 

Our model does not depend on having access to a trained language model. By dropping this requirement, we make a trade-off in performance: there are many differential equations that LC can solve exactly but our method will not.

However, the advantages of this trade-off are numerous. We do not need to train a hefty language model over tens of millions of symbolic expressions, nor do we need to exert effort to obtain this trove of training data (in the form of labelled, input-output pairs of solved integrals and differential equations). The lack of dependency grants our compact model the flexibility to adapt to new environments without training a new language model. For example, if we wish to change our set of allowable operations, we can do so at no extra cost and begin the SFL as normal.


\subsubsection{Versatility.}
Our framework was designed with differential equations in mind, but can be easily applied to a variety of other tasks with no extra effort. In addition to first and second order diffential equations, the following problems can be tackled using our framework by simply using the appropriate choices for the function $g(x, y, y', y'')$ of Equation~\ref{g_defn}:

\begin{itemize}
    \item \textbf{Integration:} Compute the antiderivative of $p(x)$ by setting $g(x, y, y', y'') = y' - p(x)$.
    \item \textbf{Functional equations:} Solve the functional equation $h(x, y) = 0$ by setting $g(x, y, y', y'') = h(x, y)$. 
    \item \textbf{Compute inverse functions:} Obtain an expression to approximate the inverse of the function $p(x)$ by setting $g(x, y, y', y'') = x - p(y)$. 
    \item \textbf{Compute roots:} Obtain a approximate numerical root of the function $p(x)$ by setting $g(x, y, y', y'') = p(y)$.
    \item \textbf{Symbolic regression:} Obtain a symbolic expression to fit a dataset of points $\{(x, y)\}$ by defining $g(x, y, y', y'')$ as $0$ (declaring it as an irrelevant aspect during training) and providing the dataset of points for fitting as the set of initial value conditions, guiding the optimization process.
\end{itemize}

\section{Limitations}

The main limitation in our model is the same that affects all approaches used for symbolic function learning: the space of all possible mathematical functions is so vast that the task of identifying the one expression best suited for a given objective function is very likely to be an NP-hard problem \cite{udrescu2020ai}. This difficulty is further compounded by the fact that neural networks are not naturally suited to approaching problems in symbolic mathematics \cite{lample2019deep}. Although our SFL discovers accurate or useful functions in many cases, it is susceptible to getting trapped in local minima and returns suboptimal results on several occasions. This problem grows more severe as the number of tree layers or allowable operations increases.




\section{Experiments}

To test our system, we performed a number of example task on a variety of problems. 

In each task, we run our system given a function $g(x, y, y', y'')$ that encodes the problem we are trying to solve. Unless stated otherwise, the SFL algorithm is run using the unary operators $U = [id, \sin, \sqrt{}]$ and the binary operator $V =[\times]$. In order to avoid illegal argument errors, we automatically compute the absolute value before entering any value as input to operations defined only on the positive half-plane, such as the square root function. We use $\delta = 1$ when computing the operator function $\textbf{p}_\delta$, so adding an explicit addition operator is not necessary.

We run our method 20 times on each task, each time using 5000 randomly generated points within the domain of the desired function $f$ to train our model. Each run proceeds for 6000 iterations and returns the function represented by the parse tree at the final iteration. We observed that more iterations than this were not needed whenever the SFL was on track to reaching convergence. During the first 1250 (= 25\%) of iterations of each run, we use the standard form of softmax function $\textbf{s}$, and switch to the discrete form $\textbf{s}'$ for the remainder of the training. This allows the model to more freely explore the function space during early stages of training, before settling down towards a single expression structure and fine-tuning constants for the remaining iterations.

The output of our system is the function produced by the SFL that obtained the lowest validation error out of all 20 runs, as per Equation~\ref{loss_eqn}.

We use a TensorFlow 2.2.0 implementation on Python 3.6, run on Microsoft Windows Server 2016 Standard OS with four Intel (R) Xeon (R) Gold 6126 CPU @ 2.60 GHz, 2594 Mhz, 12 core processors.



\subsection{Lane-Emden Differential Equations}

\begin{table*}[t]
\begin{tabular}{@{}lllll@{}}
\toprule
$m$ &
  True Expr. &
  Learned Expression &
  Error on $[1, 5]$ &
  Error on $[0.1, 10]$ \\ \midrule
$0$ &
  $1-x^2/6$ &
  $1.000 - 0.166x^2$  &
  0.000064 &
  0.00016 \\
$1$ &
  $\sin(x)/x$ &
  $0.91 - 0.85\sin(0.04x^2 + 0.01x + 0.67\sin(0.18x - 0.03) - 0.04)$ & 
  0.327 &
  5.320 \\
$2$ &
  None &
  $0.507 + 0.485\sin\left(\frac{2.0x+ 0.5}{0.3x + 1.8} + \frac{0.176x+0.007}{0.142x + 0.006} + 0.094 \right)$ & 
  0.0047 &
  0.157 \\
$3$ &
  None &
  $-0.001x^{2}-0.069x-0.407\sin(0.170x+0.016)+0.980$ & 
  0.0935 & 
  0.271 \\
$4$ &
  None &
  $-0.053+\frac{2.95\sqrt{0.7087x+0.4733} + 2.79}{0.77x^{2}+1.50x+4.50} $ &
  0.00044 &
  0.103 \\
$5$ &
  $\frac{1}{\sqrt{1+\frac{x^2}{3}}}$ &
  $1.00-\left(1.01\sin(0.18x-0.01) + 0.025\right) / \left( \frac{-1.45x-0.6}{0.1-1.6x}+0.08\right) $ &
  0.00989 &
  0.0815 \\
 \bottomrule
\end{tabular}
\caption{The Lane Emden equation, shown in Equation~\ref{LEequation}, has known solutions for $m = 0, 1,$ and $5$, but not other values of $m$. Our system is able to produce symbolic expressions to approximate solutions for all values of $m$.}
\label{LaneEmdenResults}
\end{table*}

\begin{figure*}[t]
\centering
\includegraphics[width=0.85\textwidth]{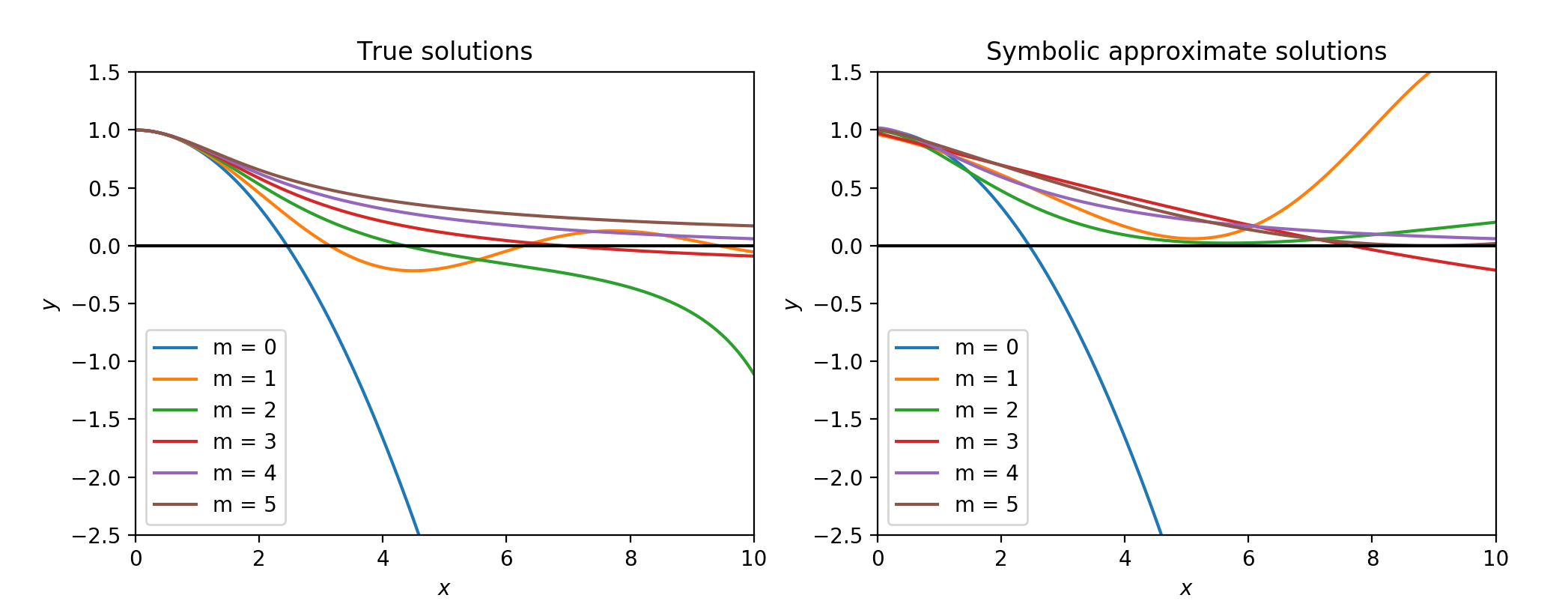} 
\caption{Graphs of Lane-Emden equations, comparing the true solutions (left) with the symbolic approximations obtained using our model (right). Note that the symbolic approximation functions are able to replicate the general behaviour of the true functions, while constrained to using only the prespecified operations in mathematical expressions of limited complexity.}
\label{LEgraphs}
\end{figure*}

The Lane-Emden equations are a family of equations relating to hydrostatic equilibrium that have applications in astrophysics \cite{adomian1995analytic}. The general form of the homogeneous Lane-Emden differential equation of order $m$ is
\begin{equation}
\label{LEequation}
   y'' + \frac{2}{x}y' + y^m = 0 
\end{equation}
where $y(0) = 1$ and $y'(0) = 0$ are initial conditions.

This family of differential equations is interesting, among other reasons, because closed-form solution to this equation are known for $m=0, 1,$ and $5$, but not for other values of $m$. It is not known if analytic solutions exist in these cases. Instead, researchers resort to computational approximations for these equations \cite{parand2010approximation}.

We ran our system on the Lane-Emden equation for values of $m$ ranging from $0$ to $5$ by setting $g(x, y, y', y'')$ as given in Equation~\ref{LEequation}. For $m=1$ onwards, we added the division operator $\div$ to our set of allowable binary operations. Our method obtained the symbolic expressions  shown in Table~\ref{LaneEmdenResults}. The error score over the interval $[a, b]$ is a measure of how close the values of the  $g$ function are to 0 on the interval when using the learned expression $f$, computed using the integral $\int_a^b \left|g(x, f(x), f'(x), f''(x))\right|^2 dx$. 

The Lane-Emden equations are meant to be applied over the semi-infinite domain $[0, \infty)$, but for computational purposes we restricted the domain of training to be within $[0.1, 4]$. Points around $x=0$ were excluded in order to avoid the singularity in Equation~\ref{LEequation}. Despite the fact that the points given in initial value conditions fall outside the domain of the training points, our system is still able to effectively abide by the given constraints. 

As seen in Table~\ref{LaneEmdenResults}, our system was able to produce competent symbolic approximations for the Lane-Emden equations even when no symbolic solution is known to exist. Aside from when $m=1$, the approximation quality was very good over the interval $[1, 5]$. Most of the additional error obtained in the extended range $[0.1, 10]$ did not come from the region $[5, 10]$ but from the minor interval $[0.1, 1]$ due to the instability near the origin.







\subsection{Integration for Non-Elementary Functions}

\begin{table*}[t]
\centering
\begin{tabular}{@{}lllll@{}}
\toprule
Name &
  Integrand, $f$ &
  Learned Expression, $\hat{F}$, for the antiderivative of $f$ &
  Interval &
  Error on Interval \\ \midrule

Bell curve &
  $ e^{-x^2}$ &
  See $\hat{B}(x)$ in text &
  [-2, 2] &
  0.00185\\
Elliptic integral &
  $ \sqrt{1-x^4}$ &
  $1.50\sin\left(0.15x^2 + 0.58x + 1.28\sqrt{|0.053x + 0.05|} + 0.05\right)$ &
  [-1, 1] &
  0.00094\\
Nested sin &
  $\sin(\sin(x))$ &
  $\sin(0.532x)(0.241x + \sin(0.541x))$ &
  [$-\pi, \pi$] &
  0.00234\\
Root of sin &
  $ \sqrt{\sin({x})}$ &
  $0.938x\sin(\sqrt{0.494x})$ &
  [0, $\pi$] &
  0.0103\\ \bottomrule
\end{tabular}
\caption{The symbolic expressions found by our method for integrals that cannot be expressed in terms of elementary functions.}
\label{integral_table}
\end{table*}

\begin{figure*}[t]
\centering
\includegraphics[width=0.95\textwidth]{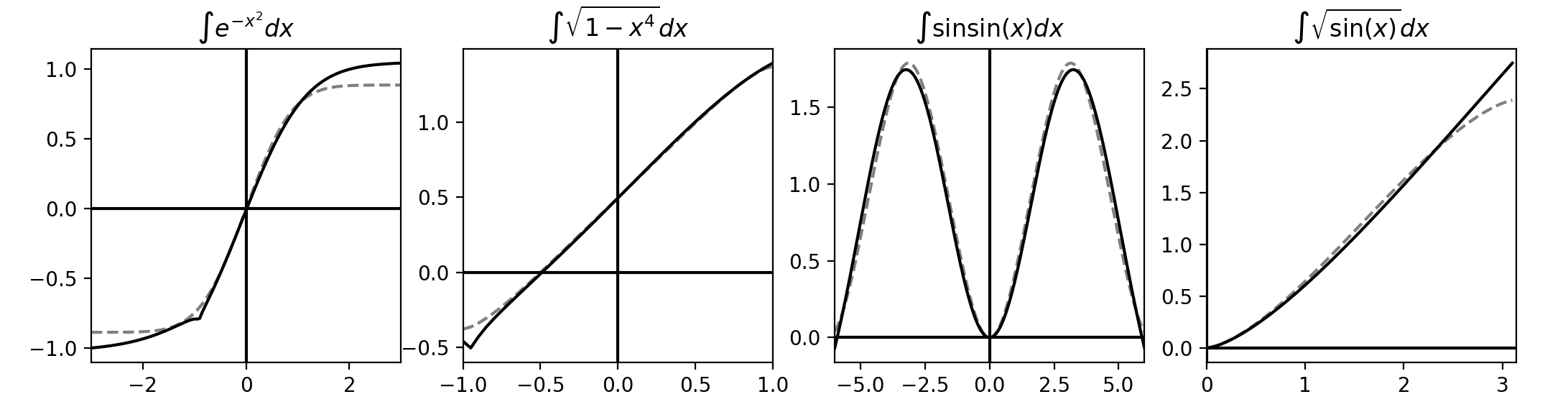} 
\caption{Each of the above graphs compares the true antiderivative (dotted line) with the symbolic approximation produced by our method (solid line) for a function whose integral is not expressible in terms of elementary functions. Constants of integration selected to line up the curves vertically for easy comparison. The symbolic expressions we found are shown in Table~\ref{integral_table}.}
\label{integral_graphs}
\end{figure*}

Many solvers already exist that are able to compute the antiderivatives of analytically integrable functions. Our method has the property of being able to generate symbolic approximations for antiderivatives for all functions $f(x)$, including those that do not have integrals that can be expressed in terms of elementary functions, by choosing $g(x,y,y')=y'-f(x)$. 

We demonstrate this ability of our method on four common examples of functions that do not have analytic antiderivatives. The results shown in Table~\ref{integral_table} use error function $Err = \frac{1}{b-a}\int_a^b \left( \int_0^xf(t)dt - \hat{F}(x)  \right)^2 dx$, where $\hat{F}$ is our approximation for the antiderivative of $f$ over $[a, b]$.

\subsubsection{The Bell Curve}

It is well known that there is no elementary function whose derivative is $e^{-x^2}$. We applied our method to approximate this integral $B(x) = \int_0^x e^{-t^2}dt$ over the set of operators $[+, \times, \sqrt, \sin]$ on the interval $[-2, 2]$ using a 3 layer SFL parse tree. 

The resulting expression obtained was 
\begin{eqnarray*}
\hat{B}(x) &=& 1.039\sin(0.969|0.982\sin(0.045x-0.105) \\
      &&-1.083\sin(0.120x+0.102) -0.107|^{0.5} \\ && +1.300\sin(1.224\sqrt{\left|0.046x+0.042\right|} \\
      &&+1.097\sin(0.451x-0.039)-0.206)-0.577) \\
      &&+0.016
\end{eqnarray*}
This is certainly a complex expression, but it fits reasonably well over the interval $[-2, 2]$, as can be seen in Figure~\ref{integral_graphs}.

Although the definition of $\hat{B}(x)$ is cumbersome, it offers insight that would not have been visible by inspection. Using this function as a starting point, we visually detected and manually applied perturbations to obtain the expression
\begin{eqnarray*}
\widehat{erf}(x) = 0.545\sin(\sqrt{\sin(0.1368x+0.0883) + 0.2120} \\
+1.300\sin(\sin(0.5162x+0.1931) -0.5716))
\end{eqnarray*}

While by no means elegant, this expression is simpler than the previous one, and also possesses a useful property: on the interval $[-1, 3]$, it is almost identically equal to the function $erf(x)=\int_{-\infty}^x e^{-t^2/2}/\sqrt{2\pi} dt$, representing the area under the normal distribution curve. The corresponding error score over this interval is
$$ \int_{-1}^3 \left(\widehat{erf}(x) - \int_{-\infty}^x \frac{e^{-t^2/2}}{\sqrt{2\pi}} dt \right)^2 dx = 0.000006464.$$

Thus, in the spirit of \cite{yerukala2015approximations}, the semi-automatically generated symbolic function $\widehat{erf}(x)$ is almost an exact match for the cumulative distribution function for the standard Gaussian distribution, and hence provides an analytic alternative to lookup tables when computing probabilities of normally distributed events.

\subsubsection{Elliptic Integral}

Another function whose integral is not expressible in terms of elementary functions is the elliptic integral
$y = \int_0^x \sqrt{1-x^4}dt$. 

We tested our framework on this integral and include the results in Table~\ref{integral_table} and Figure~\ref{integral_graphs}.

\subsubsection{Trigonmetric Integrals}

Two more functions with integrals that cannot be expressed in elementary form are $y = \sin(\sin(x))$ and $y = \sqrt{\sin(x)}$. 

We applied our framework on both of these functions and and include the results in Table~\ref{integral_table} and Figure~\ref{integral_graphs}.


\subsection{Computing Functional Inverse}

\begin{figure}[h]
\centering
\includegraphics[width=0.55\columnwidth]{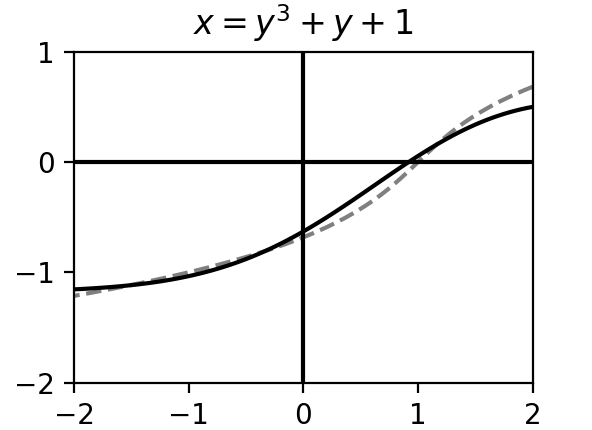} 
\caption{This graph compares the true inverse (dotted line) with the symbolic approximation produced by our method (solid line) for the cubic function $y=x^3+x+1$.}
\label{cuberoot_graph}
\end{figure}

As one final example, we use our system to produce a symbolic approximation to the inverse of  $c(x)=x^3+x+1$ in terms of the $\sin$ function. Using a 2-layer SFL, we obtain the approximate inverse function
\begin{eqnarray*}
c^{-1}(x) &\approx& 0.86\sin(0.524x +0.383\sin(0.872x-0.86) \\
&&-0.076)-0.321
\end{eqnarray*}
which we illustrate in Figure~\ref{cuberoot_graph}.

This example demonstrates how our framework can be used not only for differential equations, but also to find solutions to functional equations and compute inverse functions.



\section{Conclusions and Future Work}

We have presented a framework for using deep learning techniques to obtain symbolic solutions to differential equations, and described a neural model for learning symbolic functions that fits within the framework. Based on a number of experiments, we have shown that our method is able to effectively supply symbolic function approximations to problems that have no elementary symbolic solutions. 

Our system is open to accepting symbolic function learners other than our proposed SFL. It would be good to see how competitive symbolic regression tools would perform within our framework. Solving differential equations (and other tasks in symbolic mathematics) would be a good way to test the abilities of symbolic regression algorithms.

Our SFL has shown success in various applications as seen by our experiments, but can certainly be improved. With slight modification, our system could be extended to partial differential equations using multiple variables.

We hope this contribution will help advance the relationship between neural networks and symbolic mathematics, bringing us closer to a union of efficacy and interpretability.




\bibliography{references.bib}
\end{document}